# Transferring human emotions to robot motions using Neural Policy Style Transfer

Raul Fernandez-Fernandez [a,*], Bartek Łukawski [b], Juan G. Victores [b], Claudio Pacchierotti [c]

[a] *ISCAR, Department of Computer Architecture and Automatic, Universidad Complutense de Madrid (UCM), Spain*
[b] *Robotics Lab, Department of Systems Engineering and Automation, Universidad Carlos III de Madrid (UC3M), Spain*
[c] *CNRS, University Rennes, Inria, IRISA Rennes, France*

## ARTICLE INFO

## ABSTRACT

Neural Style Transfer (NST) was originally proposed to use feature extraction capabilities of Neural Networks as a way to perform Style Transfer with images. Pre-trained image classification architectures were selected for feature extraction, leading to new images showing the same content as the original but with a different style. In robotics, Style Transfer can be employed to transfer human motion styles to robot motions. The challenge lies in the lack of pre-trained classification architectures for robot motions that could be used for feature extraction. Neural Policy Style Transfer TD3 (NPST3) is proposed for the transfer of human motion styles to robot motions. This framework allows the same robot motion to be executed in different human-centered motion styles, such as in an "angry", "happy", "calm", or "sad" fashion. The Twin Delayed Deep Deterministic Policy Gradient (TD3) network is introduced for the generation of control policies. An autoencoder network is in charge of feature extraction for the Style Transfer step. The Style Transfer step can be performed both offline and online: offline for the autonomous executions of human-style robot motions, and online for adapting at runtime the style of e.g., a teleoperated robot. The framework is tested using two different robotic platforms: a robotic manipulator designed for telemanipulation tasks, and a humanoid robot designed for social interaction. The proposed approach was evaluated for both platforms, performing a total of 147 questionnaires asking human subjects to recognize the human motion style transferred to the robot motion for a predefined set of actions.

## 1. Introduction

The Neural Style Transfer (NST) algorithm introduced by Gatys, Ecker, and Bethge (2016) proposed a high level of abstraction for the definition of the content and style of images achieved with the VGG-19 pre-trained neural network (Simonyan & Zisserman, 2015). Features extracted through VGG-19 were used for the definition of the content and style in the context of a style transfer application. The content was defined using the layers closer to the output, responsible for extracting the higher level features of the image (e.g., people, animals, houses). The style was defined using layers closer to the input, responsible for extracting the lower level features (e.g., vivid colors, long brushstrokes). Following this concept, Gatys et al. (2016) proposed the introduction of a generic optimization algorithm to generate an image with the same low level features as the selected style image and the same high level features as the selected content image. Results were impressive (as for example combining the MonaLisa with different famous paintings: TheStarryNight, WomanwithaHat, and TheGreatWaveoffKanagawa).

The definition of a higher level of abstraction for feature extraction allowed the implementation of NST in a wide range of applications, such as in the area of motion animation (Holden, Habibie, Kusajima, & Komura, 2017). Here, the idea of NST can be introduced to generate a framework capable of automatically transforming a base animated motion which can be used to define the content (e.g., moving forward, walking in circles, moving sideways) to different animated styles (e.g., sad, happy, tired, zombie walk). As a way to perform feature extraction with motions, Holden et al. (2017) proposed the implementation of autoencoders. Fernandez-Fernandez, Victores, Gago, Estevez, and Balaguer (2022) introduced NST within discrete action spaces, where a Deep Q-Network was used for defining the control policy and performing feature extraction in the Style Transfer step.

In this paper, we propose Neural Policy Style Transfer Twin Delayed Deep Deterministic Policy Gradient (NPST3) as a way to perform Style

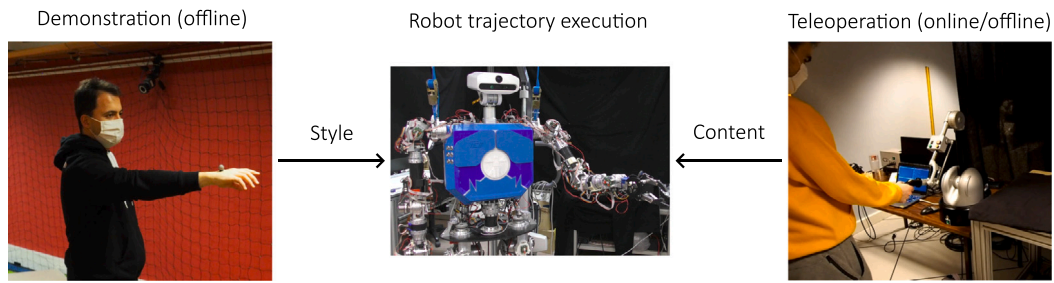

**Fig. 1.** With Neural Style Transfer, we can alter a teleoperated robotic motion (content) according to some features of a pre-recorded human demonstration (style). Such robotic motion can be carried out in e.g., an angry, happy, calm, or sad way.

Transfer with robot motions within a continuous action space. In addition to this, NPST3 allows online teleoperation while performing Style Transfer and uses autoencoders for feature extraction. This approach enables us to apply a predefined style to any robot motion, both offline, e.g., for the autonomous executions of pre-computed human-style robot motions, and online, e.g., for adapting the style of a teleoperated robot at runtime. For the generation of the robot control policies, a Twin Delayed DDPG (TD3) (Fujimoto, Hoof, & Meger, 2018) algorithm is implemented. TD3 is a promising and advanced approach to the Deep Deterministic Policy Gradient (DDPG) (Lillicrap et al., 2016) algorithm which was originally proposed as an application of Deep Reinforcement Learning (DRL) for continuous action spaces.

An application of NPST3 for the transfer of human styles (in the form of human emotions) to robot motions is proposed in the experiments section of this article. NPST3 allows the base robot motions to be generated offline or online via human teleoperation. The Style Transfer step is defined using the features extracted with the autoencoder and the TD3 network for the optimization step. The Content is defined as the high-level features that define the robot action (e.g., end-point of the trajectory), while the Style or emotion is defined using the low-level features extracted from a single human demonstration (e.g., speed, jerkiness). The TD3 algorithm allows the generation of control policies that can be executed by the robot in complex dynamic environments. Four different emotions corresponding to four different styles were selected for the experiments: angry, happy, calm, and sad. For the definition of these emotions, a single human demonstration was required. The base motions that define the content are generated using the robotic platforms and a human teleoperator. Two different robotic platform were considered: a robotic manipulator arm and a humanoid robot. The result is a robot that follows the high-level human operator instructions while introducing low-level modifications to the motions to represent the selected emotion. A schematic of this idea is depicted in Fig. 1. A preliminary version of this approach, which considered only a manipulator and reduced experimental evaluation, was proposed in Fernandez-Fernandez, Aggravi, Giordano, Victores, and Pacchierotti (2022). This improved version introduces an improved explanation of the method, an additional humanoid robotic platform and extended experimental evaluation. The main contributions of this paper are summarized in the following points:

- Proposes the NPST3 framework for performing Style Transfer with robot motions within a continuous action space.
- Allows Style Transfer to be performed both offline and online, enabling autonomous execution of human-style robot motions in dynamic environments and style adaptation of a teleoperated robot at runtime.
- Introduces TD3, a promising approach to the DDPG algorithm for continuous action spaces, within a Style Transfer framework.
- Tests the framework using two different robotic platforms, a robotic manipulator and a humanoid robot. Evaluation involved 147 questionnaires with human subjects.

- Extends the idea of Neural Style Transfer beyond images, showing that it can also be employed in robotics to transfer human motion styles to robot motions.

The goal of the proposed NPST3 framework is to allow the Style Transfer step to be performed within robot motions. This opens a new area of possible applications that find useful the decoupling of the content and style of robot motions in robot applications. Some works can be proposed using the same principle as the one introduced here for: art performances, animatronics, robot caregivers and waiters in smart city applications (Fernandez-Fernandez, Victores, Estevez, & Balaguer, 2018), craftsmanship, and in any other situation where a personalized motion is somehow important. Other works introducing NPST3, however, can include the removal of the emotion transferred by the human operator to improve the performance of the robot and remove any human bias. An implementation of our approach, including all the statistics parameters we considered, has been published in https://github.com/RaulFdzbis/NPST3.

## 2. Background and preliminaries

The success of this framework relies on the combination of two fields: Style Transfer and DRL. The first is in charge of transferring the style of one motion to other motion. The second generates the control robot policies that allow the motions to be executed in complex and dynamic environments.

### 2.1. Style transfer

Style Transfer is not a new topic within the computer animation community (Bruderlin & Williams, 1995). The first works that tried to implement this idea within motion animation introduced signal processing techniques for the definition of the style (Unuma, Anjyo, & Takeuchi, 1995). Similar recent works proposed the implementation of more advanced techniques as multilinear model design (Min, Liu, & Chai, 2010). The first works to differentiate between content and style were part of the optical character recognition area (Tenenbaum & Freeman, 1997). In robotics, Style Transfer has been bound to the introduction of emotions within robot motions; for instance, Zhou and Dragan (2018) focused on working with cost functions, and Sharma, Hildebrandt, Newman, Young, and Eskicioglu (2013) on using the Laban Effort System.

The strength of the NST algorithm proposed by Gatys et al. (2016) was the higher level of abstraction enabled with the introduction of Deep Neural Networks for the feature extraction step. In the area of motion animation, there is a lack of general motion classification neural networks that can be used as feature extractors to what Gatys did with the VGG-19 framework. As an alternative, Holden et al. (2017) implemented an autoencoder for the feature extraction step. Autoencoders are neural networks that can be trained using self-supervised learning for the encoding and decoding of inputs. The encoder layers extract the relevant features of the input to generate a compressed version of it. Then, the decoder layers take this compressed input and regenerate the original input.

With the presentation of NST by Gatys et al. (2016), a new layer of abstraction was introduced through the inclusion of the VGG-19 pre-trained classification neural network. Due to the difficulty of having a proper pre-trained motion classification neural network, Holden et al. (2017) proposed the introduction of autoencoders as an alternative. The encoder layers can therefore be used for feature extraction in the Style Transfer step. An encoder operation can be defined as $A(X) = ReLU(\Psi(X * W_0 + b_0))$, where $X$ is the input, $\Psi$ is the pooling operation posterior to the first layer, $W_0$ is the weight matrix of the encoder, $b_0$ is the layer bias, and $ReLU$ is a Rectified Linear Units (Nair & Hinton, 2010) activation.

The content can be defined as the encoder output of the autoencoder. The content loss is defined using the following equation:

$$L_{content} = \|A(C) - A(G)\| \qquad (1)$$

where $C$ and $G$ are the content and generated motions. The style can be defined as the Gram Matrix of the encoder output, and the style loss as:

$$L_{style} = \|Gm(A(S)) - Gm(A(G))\| \qquad (2)$$

where $S$ is the style motion. The total Style Transfer loss is defined as the sum of these two losses:

$$L_{st} = L_{content} + L_{style} \qquad (3)$$

### 2.2. Deep reinforcement learning

Robotic tasks entail working with continuous and high-dimensional spaces, while classical Reinforcement Learning techniques such as Q-learning (Watkins & Dayan, 1992) were originally designed for discrete action spaces. The goal of Q-learning is to find the Q-value of each action as a function of the state. This Q-value defines the action that maximizes the expected reward. In discrete action spaces, the policy can be defined as a greedy policy that chooses the action with the higher Q-value for each state. In continuous action spaces, for each state, an optimization step has to be introduced to find the action that maximizes the Q-function. If the state space is also continuous, this optimization step cannot be pre-computed, but has to be solved during the execution instead.

Lillicrap et al. (2016) proposed the introduction of DDPG to define the policy as a parametric function. The parametric policy is trained to maximize the output of the Q-function at any given state. A neural network can be designed to encode this parametric policy. As a result, the proposed DDPG introduces two different neural networks in an actor–critic architecture: one to encode the Q-function (critic), and one to encode the policy (actor). The resulting algorithm can be directly applied to find the optimal action in a continuous space. Later, an advanced and more stable version of this algorithm was developed by Fujimoto et al. (2018).

The base idea of TD3 is the same as DDPG, but some improvements were introduced to improve the convergence of the algorithm. These improvements include: the introduction of double Q-learning (Hasselt, Guez, & Silver, 2016) to avoid the overestimation problem; target policy smoothing to force similar actions to have similar values and reduce variance; and delayed policy updates to avoid updating the policy on high-error states.

### 3. Framework

The NPST3 framework is depicted in Fig. 2 and organized in four main blocks. The Style Transfer block (blue) is in charge of extracting the relevant features for the input trajectories and generating the Style Transfer loss. The constraint block (gray) defines the constraints loss that is added to the Style Transfer loss. The execution block (green) takes the Content and Generated trajectories with the total loss to generate a vector of end-effector position targets that will be executed by the robot. Finally, the robot block (red) includes the robotic platform that is in charge of executing said position targets and obtaining the generated motion trajectory.

#### 3.1. Inputs

Three 3D Cartesian motion trajectories are used as the input of the framework. These input trajectories are the Content, the Style and the Generated motions. The total length of these inputs is set to 5 s. Shorter trajectories are padded using the last value, while longer trajectories are divided in consecutive 5 s segments. The number of samples per second is fixed to 10. These settings were selected as a trade-off between motion accuracy and computational cost taking in account that, for some scenarios, they need to be generated at runtime. Input trajectories were defined using motion trajectory matrices ($C$, $S$, $G$) with an $[m, n]$ shape, where $m$ is the total number of samples, in this case 50, and $n$ is the space dimension, in this case 3.

#### 3.2. Autoencoder network: the loss network

The loss network is defined using a convolutional autoencoder. A 1D convolutional layer followed by a pooling operation layer is defined as the encoder as suggested by Holden et al. (2017). The convolutional layer uses 256 nodes and a kernel size of 5. The architecture of the decoder consists of the transpose convolutional and pooling layers of the encoder. An additional dropout layer is placed in the encoder to improve the performance. The resulting autoencoder network is used as a feature extractor for the definition of the content and style loss as in Eqs. (1) and (2), respectively.

#### 3.3. Constraints

As a measure to ensure that the framework generates feasible and acceptable motions that can be executed by the robot and show good performance, i.e., the generated trajectory should not be too fast and adapt to the robotic platform limitations, or there should no be discontinuities between trajectory executions, some additional constraints are introduced, as suggested by Holden et al. (2017). These constraints define a loss that is added to the Style Transfer loss and introduced to the TD3 algorithm for the training. Three constraints were introduced: position, end position, and velocity.

The first constraint, the position constraint, limits the position error that the Generated motion can introduce with respect to the Content motion:

$$L_p = \left\| \frac{G[t-1] - C[t-1]}{RT} \right\| \qquad (4)$$

where $t$ is the current execution step and the Robot Threshold (RT) is a handcrafted constant to normalize the values of the Cartesian workspace, assuming they are equal across all axes.

The second constraint smoothens the transition between two consecutive motion trajectories. A penalization value is introduced when the end position of the Generated motion is different than the end position of the Content motion. In motions that have been split, this forces the transition between the end of one motion and the beginning of the next to be smoother.

$$L_{ep} = \left\| \frac{G[t_n] - C[t_n]}{RT} \right\| \qquad (5)$$

In this expression, $t_n$ is the last time step. This loss is only computed once the last step of the trajectory has been reached.

The third constraint, i.e., the velocity constraint, increases the relevance of the velocity during the Style Transfer step:

$$L_v = \left\| \left( \frac{\partial G}{\partial t} - \frac{\partial S}{\partial t} \right) / RT \right\| \qquad (6)$$

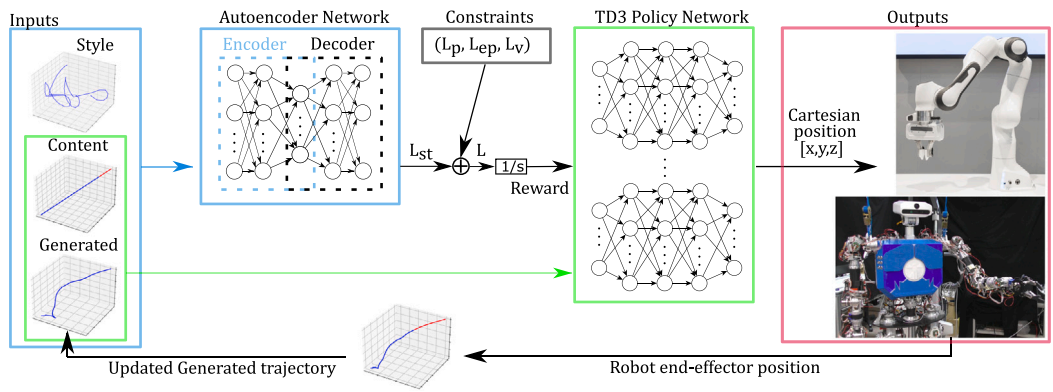

**Fig. 2.** The proposed NPST3 framework. Style, Content and Generated motion trajectories are passed to the autoencoder to compute the Style Transfer loss $L_{st}$. The constraints loss ($L_p$, $L_{ep}$, $L_v$) is added to obtain the overall loss $L$. Its inverse is used as the reward for the TD3 algorithm. In addition to this, the algorithm receives the Content and Style trajectories as input. The output is a Cartesian position vector that can be executed by the robotic platform. On each step, the motion performed by the robot is recorded and used to update the Generated motion trajectory. The content trajectory is also updated on each step with the positions defined by the user. These positions can be defined online via teleoperation or offline via a preplanned motion.

**Table 1**
Training parameters.

| Hyperparameters | Values |
| --- | --- |
| *Shared* | |
| Motion input shape | [50, 3] |
| Sample frequency | 10 [Hz] |
| Motion length | 5 [s] |
| Motion Robot Threshold (RT) | 300 [mm] |
| Optimizer | Adam (Kingma & Ba, 2015) |
| Number of styles | 4 |
| *Autoencoder* | |
| Epochs | 1000 |
| Batch size | 256 |
| *TD3 network* | |
| Action space dimensions | 3 |
| Action Range (AR) | ±0.1 * RT |
| Loss weights ($w_c, w_s, w_p, w_{ep}, w_v$) | (100, 1, 0.1, 1, 20) |
| Epochs | 2500 |
| Experience Replay size | 1000 |
| Batch size | 64 |
| Critic Learning Rate | 1e−5 |
| Actor Learning Rate | 1e−6 |
| Discount ($\gamma$) | 0.99 |
| Critic/Actor update ratio | 2 |
| Target update value (tau) | 1e−3 |
| Loss Function | Mean Squared Error |
| Initialization network values | ± 3e−3 (Uniform) |
| Policy noise | 0.002 * RT (Normal) |
| Action noise | 0.02 * RT (Normal) |

The total loss is the weighted sum of these constraints and the Style Transfer losses:

$$L = w_c L_{content} + w_s L_{style} + w_p L_p + w_{ep} L_{ep} + w_v L_v \quad (7)$$

The inverse of $L$ is used as the reward for training the execution network.

### 3.4. TD3 policy network: the execution network

The TD3 Policy network is the execution network in charge of generating the motion trajectory that minimizes the total loss. The inputs of this network are the Content and Generated motion trajectories. The network is trained to be able to accept incomplete motions. This allows these motions to be incrementally introduced to the network as in the case of the online generation of the Content trajectory. One network is trained for each of the Styles and works for any Content motion trajectory.

The TD3 architecture is divided into two different sub-networks following an actor–critic architecture (Konda & Tsitsiklis, 2000). The actor network encodes the policy of the robot, and the critic network encodes the Q-function. In the actor network, each input motion is processed separately passing through three 1D convolutional layers. The resulting outputs of these first layers are flattened, concatenated and passed to a set of four fully connected layers. The first convolutional layer has a size of 256 nodes with a kernel size of 5. This first layer is the same as the one defining the encoder. The rest of the convolutional layers have a size of 128 nodes and a kernel size of 5. In the case of the fully connected layers, the first two layers have a total of 512 nodes, the third 400, and the fourth layer 300. Batch normalization layers are defined between all the layers of the network. A ReLU activation is used for all the layers except the last one, which uses an hyperbolic tangent (tanh) activation. The output of the network is a 3D Cartesian position vector of the robot end-effector.

For the critic network, the same architecture was used with some changes. The action is introduced to the critic network as an additional input. This action is the output of the actor network and is not passed through the first convolutional layers, but directly introduced to the fully connected layers. After the first fully connected layer, the action is concatenated with the resulting output of the two other inputs. An additional fully connected layer is added before the 400 size layer. A linear activation function is used in the last layer of the network and its output is the Q-value corresponding to the input action and the current state of the content and generated motions. All parameters have been tuned during pilot experiments with the framework.

### 3.5. Outputs

The output of the NPST3 framework is a 3D Cartesian position vector used to command the end effector of a robotic platform. The motion performed is recorded and used to define the next step of the generated motion trajectory. This trajectory is the input of the next step of the framework.

### 4. Training

Similarly to the work proposed by Li et al. (2019), the CMU Graphics Lab Motion Capture Database (C.M.U. Graphics Lab, 2003) was used to train the autoencoder. This database contains a set of motions performed by volunteers while recording the position of multiple trackers placed on their bodies. Only the information from the tracker at the end of the right hand (*RFIN*) was used in this work.

For training the execution network, a random generator of linear motions was implemented for simulating the content motions. For the definition of the style motions, four different emotions were considered:

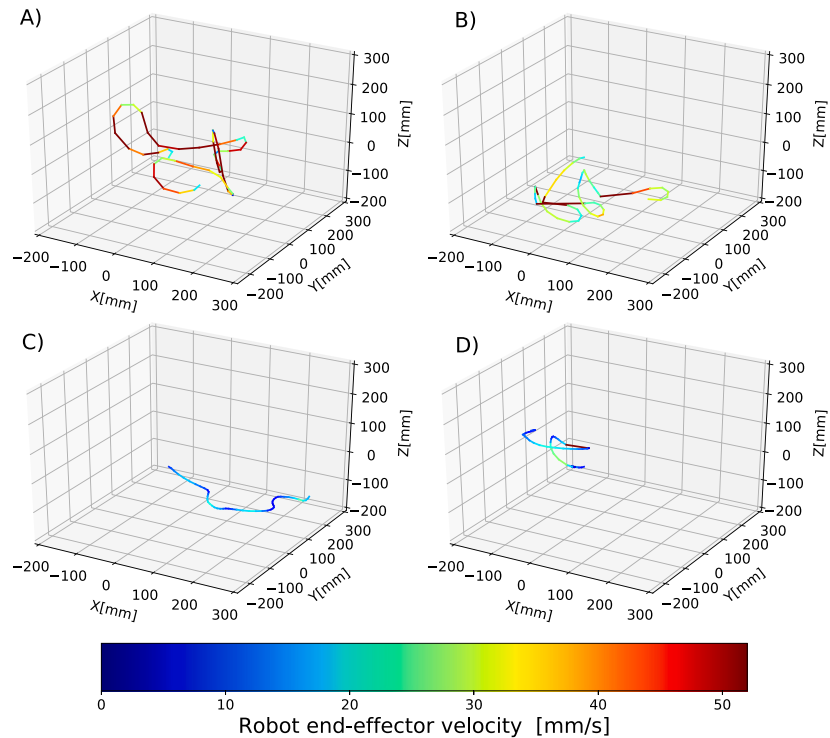

**Fig. 3.** Cartesian trajectories extracted from the human demonstrations. Depicted styles are: (A) anger/annoyance, (B) happiness/joy, (C) calm/acceptance, and (D) sadness/grief.

anger/annoyance, happiness/joy, calm/acceptance, and sadness/grief. These are polar emotions within the wheel of basic emotions as defined by Plutchik (1982). A single volunteer was asked to move his right hand freely in space, so as to convey – in any ways he wanted to – each of the four emotions mentioned above. No other restrictions or guidelines were imposed. The goal was to avoid the introduction of bias in the demonstrations.

Only one demonstration was required for each style. The position of the user's hand over time was recorded using a Vicon motion capture system. A 5-seconds-long portion of the full demonstrated motion was extracted for each style to define the style motion. One execution network was trained for each of the styles to a total of four. The same autoencoder network was used for all styles. The training parameters chosen for the autoencoder and execution networks are depicted at Table 1. These parameters were chosen after a set of preliminary experiments consisting of 50 tentative trainings, retaining the parameters best expressing the target emotions as identified by the user who carried out the demonstrations.

## 5. Experiments

An experiment involving human users and two robotic platforms, in the form of a web questionnaire, was proposed for testing the performance of the NPST3 framework. The first platform was a 7-DoF Franka Emika Panda manipulator designed for manipulation tasks. The second platform was TEO, a multi-task 30-DoF humanoid developed at Universidad Carlos III de Madrid. The manipulator was selected as a robust platform for testing the online teleoperation aspect of NPST3. The humanoid was introduced as a socially friendly platform for displaying emotions. Experiments were conducted separately for each robot. The same algorithm and trajectories were used for both robots.

### 5.1. Subjects

73 volunteers (43 males, 29 females, 1 prefer not to say; 18 to 76 years old) participated in the experiment involving the manipulator, and 74 volunteers (36 males, 37 females, 1 prefer not to say; 18 to 73 years old) participated in the experiment involving the humanoid robot. Subjects came from 10 different nationalities and 4 countries of residence.

### 5.2. Methods and task

The style motions were obtained from individual human demonstrations and are depicted in Fig. 3. For the content motion, we considered a representative simple straight line, depicted in Fig. 4 (top). These motions were introduced to the NPST3 framework as described in Section 3 to generate the four different emotion-styled motions depicted in Fig. 4 (bottom).

For simplicity, the trajectories were prepared offline, but the framework is capable of generating them at runtime, i.e., the chosen style is applied immediately to the teleoperated motion of the robot. The four resulting motions were executed on both platforms by commanding end-effector positions. In the case of the humanoid robot, the left arm (having 6 DoF) was used, keeping the rest of the body static. To avoid jerkiness, the duration of the trajectories was extended from 5 to 10 s. The initial set of 50 points was interpolated accordingly to obtain 500 points commanded every 20 ms.

The resulting robot motions were recorded through an external camera placed in front of the robot, resulting in five videos for each platform: one corresponding to the content, and four to the generated style trajectories. Four additional videos were recorded to show the manipulator robot being teleoperated at runtime with the NPST3 framework, thus demonstrating that the stylization of the robot motion can be done either offline or online. An excerpt of these videos is included as supplemental material and has been made available at https://youtu.be/WpAYniS9KOY.

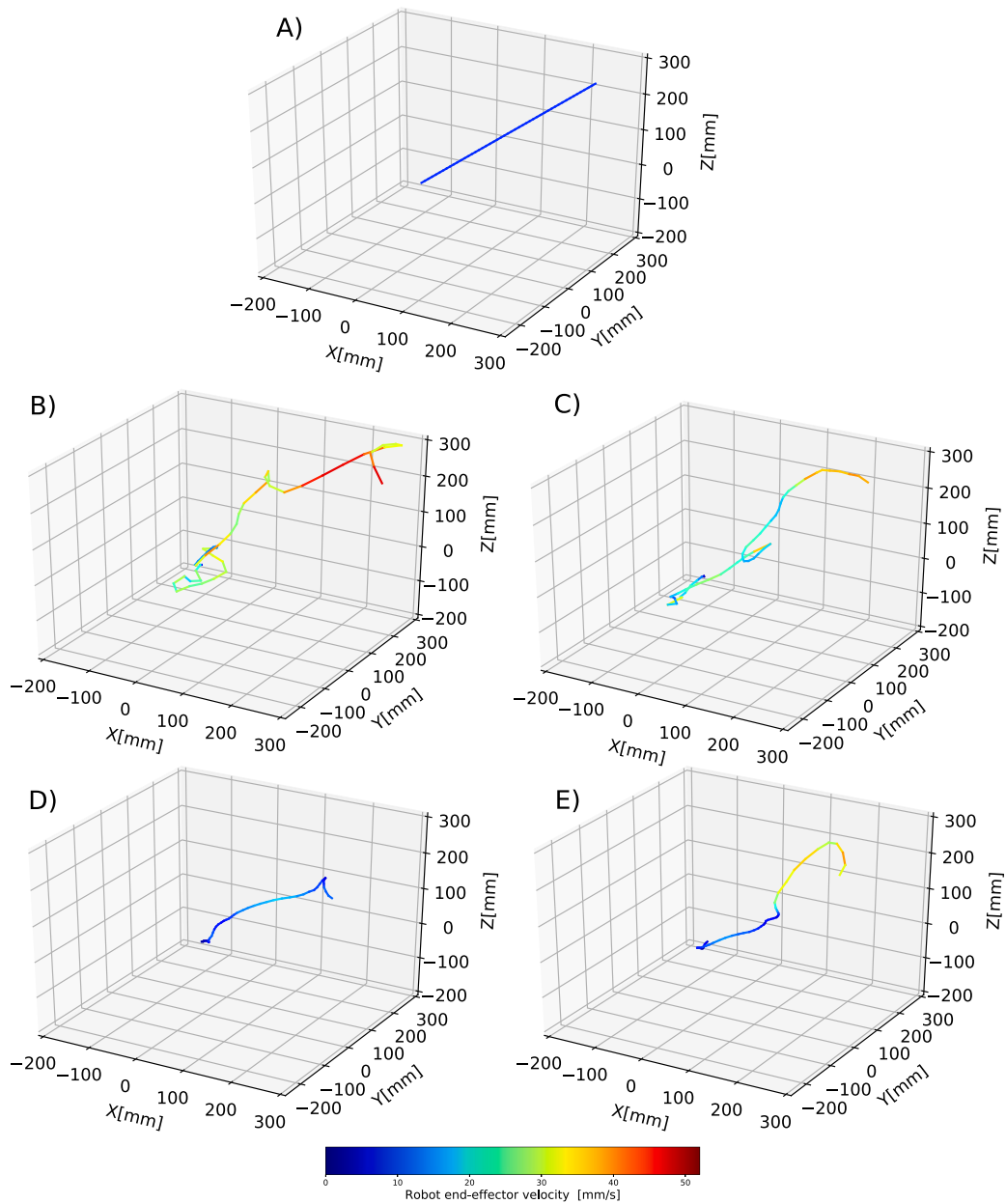

**Fig. 4.** Cartesian trajectories generated with the NPST3 algorithm. (A) depicts the Content motion. The transferred Styles are: (B) anger/annoyance, (C) happiness/joy, (D) calm/acceptance, (E) sadness/grief. The style trajectories are extracted from the movement of the right hand of a human demonstrator using a Vicon optical motion capture system.

Before the beginning of the experiment, a set of demographic questions were asked to the volunteers to gather information about their background, their predisposition to robots and the proposed application. After these questions, subjects were presented with videos showing the considered robot system in action (see Section 5.1; 73 subjects watched videos of the telemanipulator arm, 74 subjects watched videos of the humanoid robot) in the following way: one video showed the base content motion (a straight movement), four videos showed the same content motion stylized according to the four target Styles (or emotions). The videos were provided with no additional information and could be played as many times as needed.

Subjects were shown the four stylized videos twice. The first time, subjects were asked to describe, in one word, the emotion that the robotic motion elicited in them. Subjects could insert any word they preferred in an unconstrained text input. The second time, subjects were again asked to describe the emotion that this robotic motion elicited in them. However, this time they had to choose between the four emotions we used to style the content motion (anger/annoyance, happiness/joy, calm/acceptance, sadness/grief) from a dropdown menu.

### 5.3. Results

The first results obtained correspond to the demographic questions shown in Fig. 5. The goal of these questions was to learn the background and predisposition of the volunteers to like or dislike the proposed application. With respect to the background, an almost equal set of volunteers was selected among experts in robotics, volunteers with non-specialized background in robotics but having some technical background, and volunteers without any technical background. In terms of predisposition, most of the volunteers liked the idea of working with robots and considered transferring emotions to a robot an interesting topic. Some of the volunteers did not like robotics at all or considered transferring emotions to a robot a bad idea.

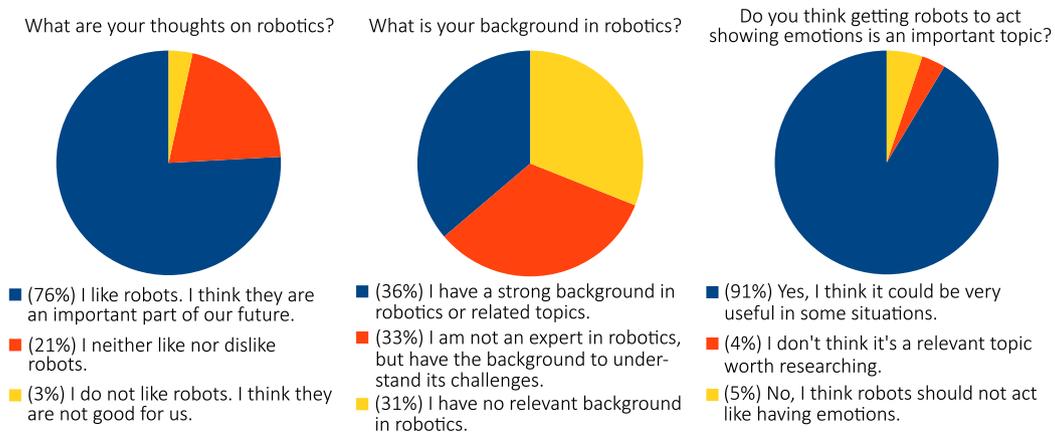

**Fig. 5.** Answers obtained from the demographic questions asked at the beginning of the experiment. The goal of these questions was to obtain information about the background in robotics of the volunteers, their predisposition to robots and the proposed application.

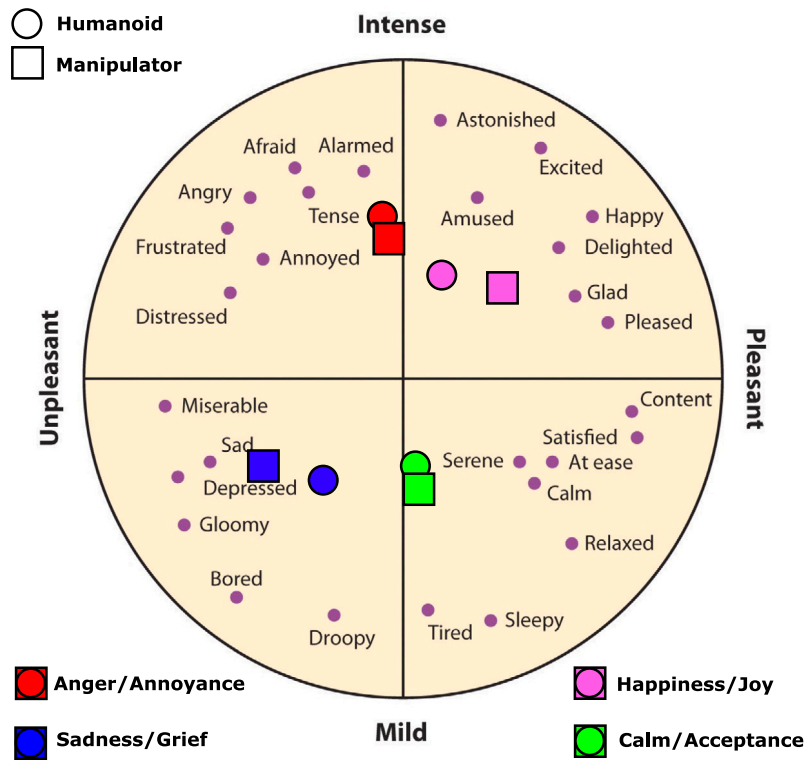

**Fig. 6.** Free text response (wheel of emotions). Given a Cartesian coordinate system growing towards "Intense" (top) and "Pleasant" (right), and its origin at the center of the wheel, a coordinate is assigned to each emotion, e.g., "pleased" is (14, 3) and "tired" is (−15, 1). The average answers of the subjects can be placed on the wheel as shown (circles and squares). Inspired by Foxcroft and Panebianco-Warrens (2015) and based on the model by Russell (1980).

Results obtained after showing the videos to the volunteers are represented in Figs. 6, 7 and 8. Results were obtained separately for each video and for each robotic platform. Figs. 6 and 7 depict the results of the first part of the questionnaire, in which subjects had to identify the elicited emotion in a free text response. In Fig. 6, the position of each emotion in the wheel of emotions, as defined by Plutchik (1982), is used to compute the average position of the identified emotions. In Fig. 7, the emotions identified by the volunteers are depicted according to their frequency. Finally, Fig. 8 provides the confusion matrices corresponding to the results of the second part, in which the volunteers subjects chose the emotion each video elicited from the set of four emotions employed to style the original content motion.

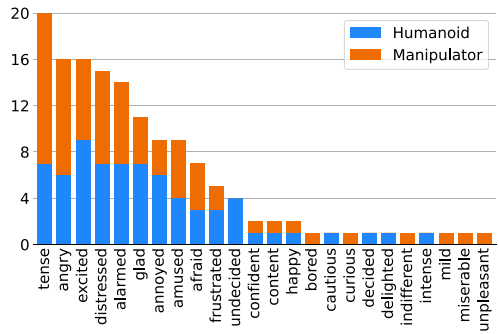
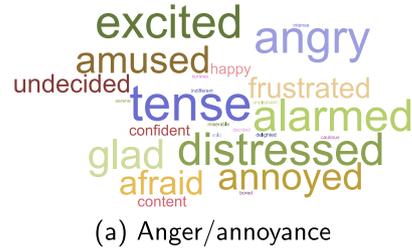

(a) Anger/annoyance

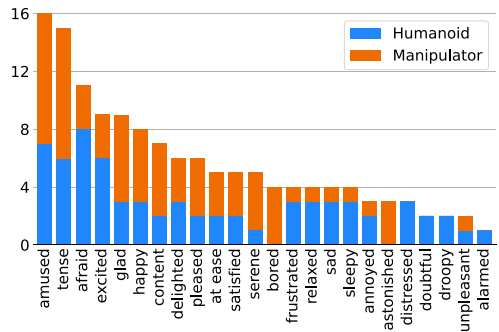
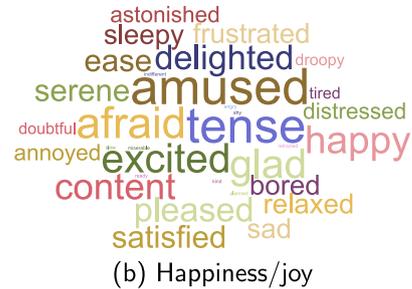

(b) Happiness/joy

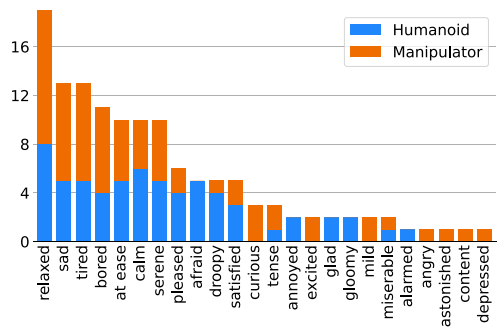
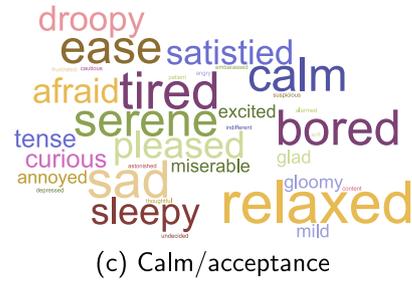

(c) Calm/acceptance

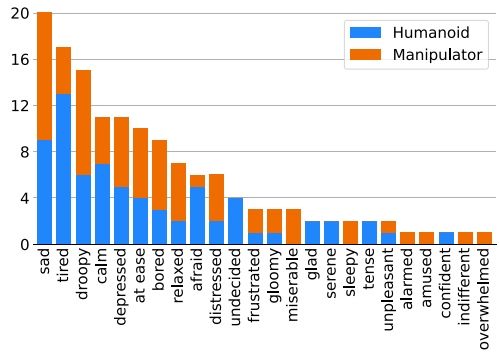
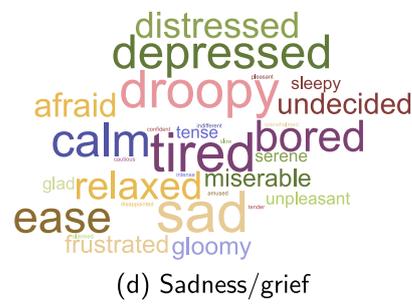

(d) Sadness/grief

**Fig. 7.** Free text response (frequency of emotions). Each pair of figures depicts the answers given by the subjects from analyzing the videos stylized with the four considered emotions. The histograms show the word frequency of said answers. This information is also represented using a word cloud (font size proportional to frequency).

Results related to moving the robot via teleoperation with the NPST3 framework are depicted in the video linked in Section 5.2. These videos were not included in the questionnaire due to the lower repeatability of teleoperated motions. This could introduce additional bias in the generated motion, making questionnaire results harder to compare.

## 6. Discussion and conclusions

In this paper, NPST3 is proposed as a way to perform Style Transfer with robot motions in continuous action spaces via DRL. The NPST3 framework is able to work with predefined motions generated offline, or online motions generated through direct human robot teleoperation.

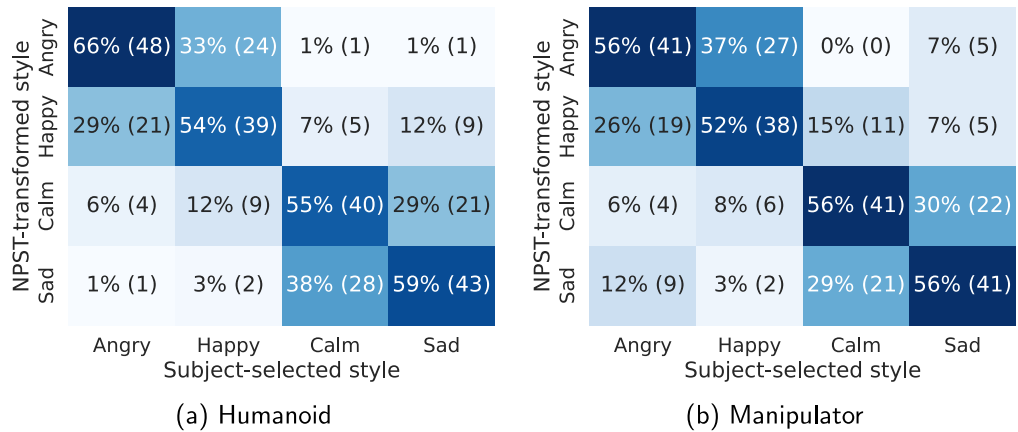

**Fig. 8.** Constrained multiple choice response. The four styles we used to alter our original content motion are represented in the rows, and the styles chosen by the subjects are represented in the columns. Each cell contains the percentage and how many times each style was selected. Correct answers are those in the diagonal cells.

Results were obtained through experiments with human subjects and two robotic platforms: a robotic manipulator arm, also used to showcase teleoperation, and a humanoid robot as a socially friendly robotic platform. For both platforms, the NPST3 framework was able to successfully transfer the style in the form of emotion to the generated trajectory. In the case of the free text response part, volunteers chose emotions similar to the one originally transferred to the robot. In the second part, in which the four considered styles were presented as mutually exclusive options, the most chosen emotion was the one originally transferred to the robot. Wrong answers mostly corresponded to pairs of emotions that share a similar intensity, i.e., anger vs. happiness and calm vs. sadness.

This work is, in our opinion, a promising starting point for introducing NST-based control in robotics. We plan to introduce the proposed approach in applications where the decoupling of content and style can be useful. For instance, a practical application would be an NST-based framework that is able to remove the emotional bias introduced by a human teleoperator in the commanded robot motion.

**Declaration of competing interest**

The authors declare that they have no known competing financial interests or personal relationships that could have appeared to influence the work reported in this paper.

**Data availability**

All the data and code has been open sourced and linked within the paper.

**Acknowledgments**

This research has been financed by ALMA, "Human Centric Algebraic Machine Learning", H2020 RIA under EU grant agreement 952091; ROBOASSET, "Sistemas robóticos inteligentes de diagnóstico y rehabilitación de terapias de miembro superior", PID2020-113508RB-I00, financed by AEI/10.13039/501100011033; "RoboCity2030-DIH-CM, Madrid Robotics Digital Innovation Hub", S2018/NMT-4331, financed by "Programas de Actividades I+D en la Comunidad de Madrid"; "iREHAB: AI-powered Robotic Personalized Rehabilitation", ISCIII-AES-2022/003041 financed by ISCIII and UE; and EU structural funds.